\def\year{2020}\relax
\documentclass[letterpaper]{article} 
\usepackage{aaai20}  
\usepackage{times}  
\usepackage{helvet} 
\usepackage{courier}  
\usepackage[hyphens]{url}  
\usepackage{graphicx} 
\usepackage{graphicx}  
\title{From Route Instructions to Landmark Graphs}

\author{\Large \textbf{Christopher Cervantes}\\ 
HERE Technologies\\ 
425 West Randolph Street\\
Chicago, Illinois 60606\\
christopher.cervantes@here.com 
}

\usepackage{latexsym}
\usepackage{amsmath}
\usepackage{amssymb}
\usepackage{algorithm2e}
\usepackage{subfig}
\usepackage{booktabs} 

\DeclareMathOperator*{\landmark}{\mathit{l}}
\DeclareMathOperator*{\jaccard}{\mathit{J}}
\DeclareMathOperator*{\words}{\mathbf{w}}
\DeclareMathOperator*{\states}{\mathbf{s}}
\DeclareMathOperator*{\actions}{\mathbf{a}}
\DeclareMathOperator*{\word}{w_i}
\DeclareMathOperator*{\state}{s_t}
\DeclareMathOperator*{\action}{a_t}
\DeclareMathOperator*{\predstates}{\mathbf{\hat{s}}}
\DeclareMathOperator*{\predactions}{\mathbf{\hat{a}}}
\DeclareMathOperator*{\predstate}{\hat{s_t}}
\DeclareMathOperator*{\predaction}{\hat{a_t}}

\usepackage{listings}
\DeclareCaptionFormat{myformat}{#1#2#3}
\begin{document}
\maketitle


\begin{abstract}
Landmarks are central to how people navigate, but most
navigation technologies do not incorporate them into their 
representations. We propose the landmark graph generation 
task (creating landmark-based spatial representations from 
natural language) and introduce a fully end-to-end neural approach
to generate these graphs. We evaluate our models on the 
SAIL route instruction dataset, as well as on a small set of 
real-world delivery instructions that we collected, and we show
that our approach yields high quality results on both our task
and the related robotic navigation task.
\end{abstract}

\section{Introduction}
As location technology has improved, there is an increased reliance on mobile and in-car apps to help people navigate in their daily lives. While these tools are well-suited to driving on established road networks, they rely on a precise geometric world representation that limits their usefulness in other navigation tasks \cite{zang2018behavioral}. 

When navigating, people use landmarks to orient themselves and define their surroundings rather than using coordinates and distance measures \cite{fellner2017turn}. Consequently, the techniques that are appropriate for automotive navigation aren't as useful when trying to find a side entrance when delivering a package, search through an unfamiliar area in an emergency, or locate a building in parts of the developing world where addressing is not well-defined. In all such cases, representing a route's landmarks relative to one another can be more useful than coordinate-based localization.

\begin{figure}[h]
	\centering
	\includegraphics[scale=0.6]{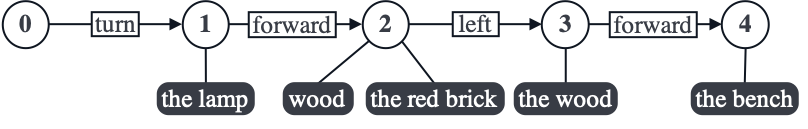}
	\caption{Landmark graph for: ``Go away from the lamp to the intersection of the red brick and wood. Take a left onto the wood. Position one is one section down at the bench."}
	\label{fig:graph_ex}
\end{figure}

We propose a method for automatically extracting landmark and relation information from route instructions. Specifically, given a natural language route instruction (e.g. ``Go away from the lamp to the intersection..."), our goal is to produce a \emph{landmark graph} such as that shown in Figure~\ref{fig:graph_ex}, where nodes represent semantically meaningful locations (landmarks or decision points) and where edges indicate spatial information. 

Route instructions allow people to communicate complex spatial information, which they do in part by focusing on the salient landmarks needed to define and travel through an environment. When interpreted, however, route instructions require a person to correctly reconstruct the spatial information into a mental model. This can be challenging because the instruction writer and reader may talk about space differently, which can lead to divergent mental models. Our approach can help to bridge this gap. By extracting spatial information from instructions, we can create consistent, landmark-based spatial representations -- landmark graphs -- that can be useful in navigation tasks. Moreover, our approach's focus on extracting complete spatial representations may also be useful for similar tasks like robotic navigation, where spatial information consists of landmarks and traversal actions.

The main technical contribution of our approach is the joint prediction of landmark spans and actions from route instructions through the incorporation of an additional attention mechanism into an encoder-decoder model. This joint model yields significant improvements over the action-only baseline when evaluating action prediction performance. We also consider the introduction of the landmark graph generation task -- that is, creating complete spatial representations from natural language -- to be an important contribution, and show that our joint approach predicts landmark graphs with a high degree of similarity to the ground truth.

\section{Related Work}
In traditional geometric representations of space, a location is an abstraction independent from a referent. A latitude / longitude pair, for example, is meant to define a fixed point on the globe. While these representations can be useful, landmarks play a much more central role in navigation \cite{fellner2017turn}. In this context, a landmark is a physical object \footnote{Places defined by physical objects may also be landmarks; e.g. ``the end of the hall" defines a component of an object.} that can be used as a reference point in mental representations of space \cite{kai2016landmarks}. Seen in this way, landmarks are not \emph{at} a location, but \emph{define} a location. This is the core idea behind our location graphs: locations are defined not by coordinates but by their relationships to landmarks.

Similar landmark-based spatial representations have been proposed, from topological maps -- graphs where nodes represent places and edges denote traversability or connection \cite{landsiedel2017review} -- to navigation graphs -- where nodes represent landmarks or decision points and edges represent paths of travel \cite{yang2015generation}. While these kinds of representations are often studied in the context of indoor navigation tools \cite{tsetsos2006semantically,yang2015generation}, they are more useful generally, particularly as an intermediate step between natural language and the navigation task \cite{fellner2017turn,zang2018behavioral}. 

Natural language route instructions are a common way to communicate spatial information at pedestrian resolutions (e.g. ``turn left at the fountain and continue down the hall"). Significant work has been done to understand and extract spatial meaning from these instructions in the robotic navigation domain \cite{macmahon2006walk,kollar2010toward,chen2011learning,mei2016listen,duvallet2016inferring,chen2019touchdown,anderson2018vision}. In that setting, models are often trained to guide an autonomous agent through a virtual environment using unstructured language; the input to the system is a route instruction, the output a sequence of actions the agent must take, and the measure of success is whether the agent reached the goal destination. 

This line of inquiry is similar to our own in that spatial information is extracted from natural language, but the goals differ. In the robotic navigation literature, guiding an autonomous agent can be thought of as a search problem; the world is represented as a grid or a graph, and the system must find a path from start to finish. In our setting, we want to represent a real-world environment by constructing a landmark graph. Robotic navigation can thus consider the spatial representation to be latent; as long as the agent reaches the goal, the model is successful. Our task, however, is concerned explicitly with this spatial representation.

In practice, this means that while robotic navigation approaches assume the agent will have access to spatial information at inference time through images \cite{chen2019touchdown,anderson2018vision} or object labels \cite{macmahon2006walk,chen2011learning,mei2016listen}, our approach treats this information as part of the output. 

Despite these differences, we borrow an important concept from the robotic navigation literature: the decomposition of the navigation task into \emph{states} and \emph{actions} \cite{chen2011learning,mei2016listen}. In their framing, navigation is the process by which an agent takes actions (moving or turning) traversing from one decision point to another. Each decision point is predefined (e.g. grid intersections) and associated with a (possibly empty) set of world states describing nearby actions. 

Informed by the landmark literature, our approach assumes no predefined decision points; they exist only in relation to nearby landmarks and the spatial relations (indicated by actions) to other decision points. This reframing allows us to adapt approaches from the robotic navigation literature to the task of generating landmark graphs.

We consider the graph's primary use to be as a navigational aid in real-world environments where a-priori knowledge about the space described in a route instruction is unavailable. Without this knowledge, however, our approach must identify which parts of the sentences refer to landmarks. Finding these \emph{landmark spans} is similar to mention detection for coreference resolution \cite{peng2015joint,lee2017end} or referring expression detection for reference resolution (grounding) \cite{krishnamurthy2013jointly,kong2014you,kennington2015simple,plummer2015flickr30k,plummer2017phrase}. In such work, relevant noun phrases must be found as part of a larger task: clustering mentions or linking expressions to image referents. Similarly, our approach must identify landmark spans to define the space described by a route instruction. 

By combing landmark spans with actions linking decision points, we develop the first fully end-to-end mechanism for generating landmark-based spatial representations from natural language route instructions.

\begin{figure*}[tb!]
	\centering
	\includegraphics[scale=1.15]{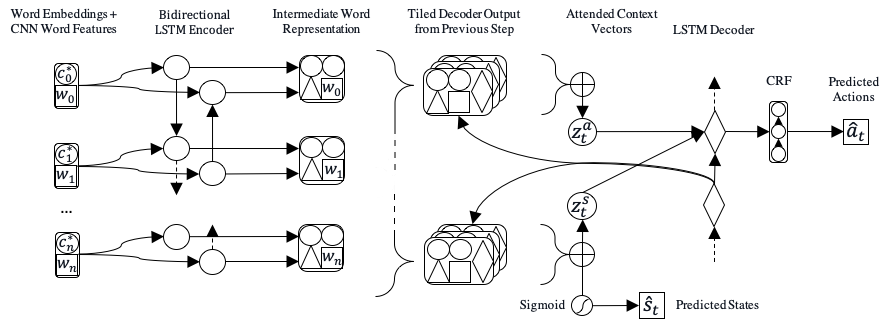}
	\caption{Full system architecture for predicting sequences of actions and states from a given sentence; word and character embeddings are passed to an encoder; outputs are concatenated with word features and combined with tiled decoder outputs; attention mechanism over word representations for each time step produces context vectors; learned attention parameters predict states; context vectors are passed to a decoder to predict actions}
	\label{fig:network_arch}
\end{figure*}

\section{Task}
We consider the task of landmark graph generation from natural language route instructions: free-form imperative statements (single or multi-sentence) that can be used to guide an agent through an environment. Landmark graphs are composed of two types of nodes -- decision points and landmarks -- and two types of edges -- between decision points and landmarks (indicating nearness) and directed from one decision point to another with an action label (indicating the path of traversal and thus the spatial relation between points). 

The task can be considered a form of summarization by which action sequence $\actions$ and world state (landmark) sequence $\states$ are extracted from instruction $\words$ ($w_i \in \words$). These same-length sequences ($|\actions| = |\states|$) represent a path of traversal; each step (decision point) in the path has nearby landmarks ($\state$) and an action indicating how the current step was reached ($\action$). 

We consider nine actions: \emph{stand}, \emph{forward}, \emph{left}, \emph{right}, \emph{ascend}, \emph{descend}, \emph{turn}, \emph{move}, and \emph{STOP} ($a_t \in \{s, f, l, r, a, d, t, m, \langle STOP \rangle \}$). In addition to the \emph{forward}, \emph{right}, \emph{left}, \emph{stand}\footnote{Our \emph{stand} action, like \texttt{verify} in the SAIL dataset, is used as a representational tool associating landmarks with decision points (i.e. when a sentence describes nearby landmark without directing any movement); unlike \texttt{verify}, \emph{stand} is only used in these cases, rather than as an anchor for landmarks}, and \emph{STOP} actions that are standard in the literature, we also include \emph{ascend} and \emph{descend} to account for three dimensional movement (e.g. climbing stairs). We also include the ambiguous \emph{turn} and \emph{move} actions for cases where the exact movement is unclear from the text.

Landmarks are noun phrases that describe a physical object useful in the navigation task. World state $\state$ represents the possibly empty set of text spans in the input sentence $\words$ that refer to landmarks near the $t^{th}$ decision point. We represent $\state$ as a binary vector of the same length as the input sequence ($\state \in \{0, 1\}^n$), where $\word$ is a token that refers to a landmark near step $t$ \emph{iff} $s_{ti}=1$.

\subsection{Graph Construction} 
Given action and state sequences $\actions$ and $\states$, the landmark graph construction process is shown in Algorithm~\ref{algo:graph_construction}, where $T$ refers to the length of the sequence.

\begin{lstlisting}[caption={Landmark Graph Construction from \\Actions ($a_t \in \actions$) and States ($s_t \in \states$)},captionpos=bc,label={algo:graph_construction},mathescape=True]
Add root node: $n_0$
For t=1 to T:
  Add new step node: $n_t$
  Add labeled edge: ($n_{t-1}$, $a_t$, $n_t$)
  For each landmark $l$ specified in $s_t$:
    If $l$ is not in the graph:
      Add new landmark node: $l$
    Add edge: ($l$, $n_t$)
\end{lstlisting}

As defined here, landmark graphs contain only landmark proximity information (i.e. an unlabeled edge between a landmark and decision point indicate nearness). While the rest of this paper assumes that landmark graphs will take this form, it is conceptually trivial to label these links with relative directions (e.g. \texttt{in\_front\_of}) to increase the expressivity of the graph. 

Though landmark graphs are similar to the graphical representation introduced in Chen and Mooney \shortcite{chen2011learning}, our representation -- informed by the landmark and navigation literature -- communicates locations with nodes and spatial relations with edges, resulting in a cleaner, more intuitive connection between the described space and the navigation task.

\section{Approach}
The central intuition behind our approach is that in order to understand a route instruction, a sentence must be parsed multiple times, focusing on different phrases on each pass to construct a spatial representation. A route is thereby decomposed into decision points; at each, our models ask \emph{where am I} -- which landmarks are near the current position -- and \emph{how did I get here} -- which action was taken to traverse from the previous step to the current -- all while keeping track of where it's been. To do this, the model encodes the sentence and decodes a sequence of actions, at each step attending to the parts of the sentence that inform the action and separately attending to the parts of the sentence that describe nearby landmarks. 

Specifically, we apply an encoder-decoder approach to predict action sequence $\predactions$ and state sequence $\predstates$ from route instruction $\words$, where the length of the output sequences are bounded by the prediction of the \emph{STOP} action or by reaching the maximum sequence length. Given $\predactions$ and $\predstates$, landmark graphs are constructed following Algorithm~\ref{algo:graph_construction}.

The full network architecture is shown in Figure~\ref{fig:network_arch}

\subsection{Encoder} 
We represent words with the combination of learned character-level features and dense embeddings. Specifically, each character $c^j_i$ in word $\word$ is represented as a one-hot-vector combined with explicit features (e.g. \texttt{is\_digit}, \texttt{is\_alpha}). The word's character representations are passed to a convolutional neural network, where filters are swept over character groupings and max-pooling is applied to the filter outputs to produce fixed-length character-level features for the word: $c^*_i$. These features are then concatenated with the pre-trained Word2Vec embedding \cite{mikolov2013distributed} for $\word$. These word representations are then passed to a bidirectional long-term-short-term-memory network (LSTM) \cite{hochreiter1997long} to encode the word in its context.

\subsection{Decoder} 
Following Mei et al. \shortcite{mei2016listen}, we produce an action context vector $z^a_t$ for each time step, first by combination of forward and backward encoder outputs $e_i := [e^{fw}_i, e^{bw}_i]$, the word embedding $\word$, and the decoder hidden state from the previous time step $d_{t-1}$ (tiled across input words). We also concatenate simple explicit word features $\phi_i$ to this representation (e.g. part-of-speech tags). This combined representation is then attended over, such that the context vector $z^a_t = \sum_i \alpha^a_{ti}[w_i, e_i, \phi_i]$, where attention weights $\alpha^a_{ti}$ are learned according to the following:
\begin{equation}
\begin{aligned}
\alpha^a_{ti} &= \dfrac{\exp(\beta^a_{ti})}{\sum_i \exp(\beta^a_{ti})} \\
\beta^a_{ti} &= v^a \text{tanh}(W^a d_{t-1} + U^a w_i + V^a [e_i, \phi_i])
\end{aligned}
\end{equation}
where $v^a$, $W^a$, $U^a$, and $V^a$ are learned parameters. Unlike Mei et al. \shortcite{mei2016listen}, however, our system does not concatenate a ground truth world state to this action context vector. Instead, we learn a world state context vector $z^s_t$ using the technique defined above. The combined vector $z_t := [z^a_t, z^s_t]$ is then passed to a unidirectional LSTM decoder.

\subsection{Prediction} 
The decoder output is passed to a linear chain conditional random field layer (CRF) which finds the best action, $\predaction$, based both on the label scores and on the predictions for previous time steps. When predicting a world state, $\predstate$, we pass the vector $[\beta^s_{t0}, \beta^s_{t1}, ... \beta^s_{tn}]$ to a sigmoid function and consider all positive values as indicative that a word describes a landmark; these words are then grouped naively into spans. 

Though learning $\predstate$ directly from $\beta^s_{ti}$ is a relatively simple modification, the conceptual novelty is important to explore. In using attention to create action context vector $z^a_t$, the system is learning $\beta^a_{ti}$ to determine whether $\word$ is helpful to predict action $\hat{a}_t$. We apply this same insight to the world state representation. At each time step, we assume some number of words ($\geq0$) refer to landmarks near that step in the path; $\beta^s_{ti}$ thus serves as a score for whether the $\word$ refers to a landmark for step $t$.

We learn separate context vectors $z^a_t$ and $z^s_t$ specifically because while $\beta^a_{ti}$ and $\beta^s_{ti}$ are learning to attend to parts of the input sentence based on the current position in the predicted path, their goals are different; words that indicate which action to take (e.g. ``then turn left") are not the same words that indicate landmarks (e.g. ``the corner of the house"). Both vectors are necessary for the decoder, however, as knowing the previous action and nearby landmarks is necessary to understand the current location.

\subsection{Training} 
\label{subsec:training}
During training, we use negative log likelihood loss for actions: $\mathcal{L}^a = -\sum \text{log} P(\predactions | \mathbf{z})$. The CRF probability for an action given a context vector is given in Equation~\ref{eq:crf_proba}, where $\mu_t^a$ is the unary score for context vector $z_t$ taking action $\action$, the probability that action $a_{t+1}$ follows action $\action$ is $\theta_{\action, a_{t+1}}$, and $\zeta$ is a normalization term (the sum of combinations over all possible actions at each time step). 
\begin{flalign}
P(\predactions | \mathbf{z}) &= \frac{1}{\zeta} \exp \left( \sum_t^T \mu_t^a + \sum_t^{T-1} \theta_{\action, a_{t+1}} \right) &&
\label{eq:crf_proba}
\end{flalign}
For states we use sigmoid cross entropy loss shown in Equation~\ref{eq:state_loss}, where $s_{ti}$ refers to the binary world state label for word $\word$ at time step $t$
\begin{multline}
\mathcal{L}^s = -\sum_t\sum_i \max(\beta^s_{ti}, 0) - \beta^s_{ti} s_{ti} + \\
\log(1 + \exp(-|\beta^s_{ti}|))
\label{eq:state_loss}
\end{multline}
We train our model using joint loss $\mathcal{L} = \mathcal{L}^a + \mathcal{L}^s$.

\section{Experiments}
Ideally, we would evaluate our system by measuring to what extent generated landmark graphs were helpful in real-world human navigation tasks. In practice, however, we must focus on whether the generated graphs contain the same information as those of the ground truth, either through the constituent elements -- actions and states -- or by comparing the complete graphs.

\subsection{Data}
In our experiments, we train and evaluate models with two datasets. The SAIL route instruction dataset \cite{macmahon2006walk} contains three maps and natural language route instructions annotated with actions\footnote{SAIL's \texttt{travel}, \texttt{turn left}, \texttt{turn right}, and \texttt{NULL} actions can be trivially transformed to our \emph{forward}, \emph{left}, \emph{right}, and \emph{stand}; \texttt{travel(step: n)} is interpreted as $n$ \emph{forward} actions}, states\footnote{SAIL landmark entities are associated with a \texttt{verify} action; we attach these landmarks directly to decision points.}, and path coordinates. Since our approach operates over surface realizations rather than the fixed set of entities in SAIL (e.g. ``the red brick" or ``the brick alley" instead of \texttt{BRICK HALLWAY} ), we augment their annotations with a by-sentence mapping from entities to surface strings using a heuristic approach reviewed by annotators. 

While the SAIL dataset is the most appropriate publicly available dataset for our task, the instructions describe simple virtual worlds. Since we are motivated by real-world environments such as those encountered by package handlers, we collected a toy dataset of route instructions (82 paths; 188 sentences) that begin at some referential point (a street near an address) and end at a final delivery location. We refer to this as our \emph{Delivery} dataset.

Despite being significantly smaller than SAIL ($12\%$ as many routes; $6\%$ as many sentences), our Delivery dataset has about as large of a vocabulary (Delivery: 481; SAIL: 587) and much longer sentences (Delivery: 17.1; SAIL: 7.8). Our Delivery dataset is also more referential; where SAIL contains 527 landmark surface realizations (0.8 per sentence; 3.7 per route), our dataset contains 324 (2.5 per sentence; 5.6 per route). While this dataset is significantly smaller than SAIL (which is itself much smaller than corpora traditionally used in neural systems), we believe these experiments can provide important insights into our approach's applicability to real-world environments.

\subsection{Experimental Setup} 
We use convolutional filters of size 2, 4, 8, and 16 over characters to produce character-level word feature vectors of length 16, pretrained Word2Vec embeddings of size 300, encoder hidden states of width 200, 20\% dropout on encoder inputs and 50\% dropout on encoder outputs. The decoder hidden state width is 256, and a batch size of 1. We train for 50 epochs using an exponentially decaying learning rate (initial: 0.001; rate: 0.99; steps: 1000) in conjunction with the Adam optimizer \cite{kingma2014adam}. 

In the following sections, all reported quantitative results are three-fold averages of models trained on train + development data and evaluated on test data (where the train/dev/test split is 80/10/10). The qualitative examples shown in Figures \ref{fig:sail_example} and \ref{fig:lmd_example} are predictions on dev. data made by models trained on the corresponding train fold. Similarly, the hyperparameters were tuned on one of the SAIL train folds and evaluated on the corresponding dev. fold.

\begin{table}[tb]
\centering
\begin{tabular}{l | l l | l}
\toprule
& \emph{MLA} & \emph{Ours (SAIL)} & \emph{Ours (Delivery)} \\
\midrule
Acc. & 68.3\% & 90.6\% & 68.9\% \\
\bottomrule
\end{tabular}
\caption{Action prediction accuracy}
\label{tbl:action_acc}
\end{table}

\begin{table}[tb]
\centering
\begin{tabular}{l l c c c c}
\toprule
\multicolumn{2}{c}{Distance} & 0 & 1 & 2 & 3\\
\midrule
\multicolumn{6}{l}{\emph{MLA}}\\
& Sent. & 49.9\% & 55.8\% & 77.5\% & 85.4\% \\
& Route & 14.6\% & 22.1\% & 27.7\% & 33.8\% \\
\multicolumn{6}{l}{\emph{Ours (SAIL)}}\\
& Sent. & 88.6\% & 94.7\% & 99.3\% & 99.7\% \\
& Route & 50.2\% & 56.8\% & 64.8\% & 66.7\% \\
\midrule
\multicolumn{6}{l}{\emph{Ours (Delivery)}}\\
& Sent. & 60.1\% & 88.8\% & 98.6\% & 100.0\% \\
& Route & 37.0\% & 70.4\% & 85.2\% & 85.2\% \\
\bottomrule
\end{tabular}
\caption{Goal position accuracy as a function of distance for single-sentence and full route instruction action sequences}
\label{tbl:dist_eval}
\end{table}

\section{Results}
We evaluate our approach in three ways: by the predicted actions, landmark spans, and complete landmark graphs. As in previous work, we treat sentences independently from one another. Where route-level measures are shown, the sentence-level predictions were combined linearly, where all root nodes except for the first are dropped, and any \emph{stand} action at the beginning of a sentence-level graph is replaced with \emph{move} (as such an action indicates an uncertain anchoring at a landmark).

Using these measures, we evaluate the performance of two separate models trained on the SAIL and Delivery datasets, respectively. In order to compare our approach to similar systems, we treat the Multi-Level Aligner model (MLA) from Mei et al. \shortcite{mei2016listen} as our baseline, training and evaluating their system\footnote{Our MLA baseline results are actions and states produced by the publicly available code trained for 50 epochs.} on the same folds as our SAIL model. We focus specifically on MLA -- rather than more recent approaches like \cite{anderson2018vision} and \cite{chen2019touchdown} -- because it is the approach on which ours is based. The differences in performance can thus be seen as an approximate measure of the usefulness in enabling the system to learn to find landmark spans, rather than being given a simple world state representation at inference time.

\subsection{Actions}
We borrow two action prediction evaluation measures from the robotic navigation literature: whether the predicted action was correct at each step (accuracy) and whether an agent following those actions arrived as a goal position given a distance threshold (distance). 

While the SAIL data contains ground truth path coordinates, our Delivery data -- and any similarly constructed real-world dataset -- does not. We therefore must handle ambiguous actions: \emph{move} $\rightarrow$ \emph{forward}, and \emph{turn} has a 50/50 chance of being \emph{right} or \emph{left}. This randomness should have minimal effects given our three-fold validation and the rarity of \emph{turn} actions, but this means the distance measure on the Delivery data is less consistent. 

It's important to note that our measures are distinct from previous work in two ways. First, we entirely disregard orientation, as our simplified landmark graph representation only considers a landmark's nearness, rather than its relative direction. Second, we capture cases where a final position is close to but not matching a goal position with Euclidean distance, rather than the number of intersections in the SAIL grid. While the overall effect of this discrepancy should be minimal, it makes it difficult to compare our work directly with previous papers like \cite{artzi2013weakly} and \cite{andreas2015alignment}.

The action accuracy and distance results are shown in Tables \ref{tbl:action_acc} and \ref{tbl:dist_eval}, respectively.

Our SAIL model significantly outperforms the MLA baseline (+23.3\% action accuracy; +38.7\% single sentence goal accuracy; +35.6\% route goal accuracy) on which it is based, suggesting that jointly learning to identify nearby landmarks helps the model predict actions. 

The performance of our Delivery model is more modest, suggesting both the increased difficulty of that setting and of training a model on so few examples. Our Delivery model does exhibit the same accuracy increase as a function of distance as both our SAIL model and the MLA baseline, suggesting that the model is still learning to capture spatial relationships in this setting (particularly when permitting a distance threshold of 1).

\subsection{Landmarks}
We evaluate the performance of our models on the landmark span detection task in two ways. In the more traditional measure, we compare the predicted landmark spans with the ground truth using precision, recall, and F1, where a predicted span is correct \emph{iff} it matches the ground truth exactly. Our more permissive measure compares the tokens of the predicted and ground truth spans using the Jaccard index: $\jaccard({\landmark}^p, {\landmark}^g) = \vert {\landmark}^p \cap {\landmark}^g \vert / \vert {\landmark}^p \cup {\landmark}^g \vert$, where $\landmark$ refers to the set of tokens in a span ($w \in \landmark$).

We evaluate the ability of our models to identify the landmark spans across steps, sentences, and routes. These results are shown in Table~\ref{tbl:lndmrk_sim}. Since the MLA baseline does not predict landmark spans, no results are shown.

\begin{table}[t]
\centering
\begin{tabular}{llc c c c}
\toprule
	& & $\jaccard$ & P & R & F1\\
	\midrule
\multicolumn{3}{l}{\emph{Ours (SAIL)}} & & & \\
& Step & 71.6\% & 0.0\% & 0.0\% & 0.0\% \\
& Sent. & 68.9\% & 30.7\% & 31.8\% & 31.0\% \\
& Route & 63.7\% & 44.8\% & 46.2\% & 44.6\% \\
\midrule
\multicolumn{3}{l}{\emph{Ours (Delivery)}} & & & \\
& Step & 39.6\% & 0.0\% & 0.0\% & 0.0\% \\
& Sent. & 40.3\% & 6.5\% & 6.7\% & 6.3\% \\
& Route & 39.6\% & 8.3\% & 10.2\% & 9.0\% \\
\bottomrule
\end{tabular}
\caption{Landmark span prediction performance measured across steps (correct spans at the right step), sentences (correct spans for a sentence), and routes (correct spans for the route instruction)}
\label{tbl:lndmrk_sim}
\end{table}

\begin{table}[tb]
\centering
\begin{tabular}{l|cc|cc}
\toprule
	& \multicolumn{2}{c|}{Sent. (sim$\vert$$\text{sim}^{\ell}$)} & \multicolumn{2}{c}{Route (sim$\vert$$\text{sim}^{\ell}$)} \\
\midrule
\emph{MLA} & \multicolumn{2}{c|}{76.6\%} & \multicolumn{2}{c}{65.9\%} \\
\emph{Ours (SAIL)} & 91.1\% & 92.9\% & 83.1\% & 88.0\% \\
\midrule
\emph{Ours (Delivery)} & 67.6\% &73.3\% & 57.0\% & 63.3\% \\
\bottomrule
\end{tabular}
\caption{Graph similarity for sentences and routes using the strict (sim) and relaxed ($\text{sim}^{\ell}$) measures}
\label{tbl:graph_sim}
\end{table}

For both sentences and routes, our SAIL model finds approximately the correct range of tokens that refer to landmarks (evidenced by the high Jaccard index) and the exact span (shown by the F1 score). While these scores are poor in comparison to modern methods for mention detection\footnote{
Peng et al. \shortcite{peng2015joint} reports mention detection scores in the 70-90\% range}, they show that a fully end-to-end approach for capturing both spatial relations and landmark information in one system is beginning to yield positive results.

Though our Delivery model behaves similarly to our SAIL model, the performance is significantly worse (likely due to the small dataset). It's also worth noting that for both models the step-level F1 is approximately 0 despite a similar Jaccard index for steps and sentences. This is likely because when a landmark span is predicted it overlaps meaningfully with the ground truth, but the vast majority of steps have no predicted span and thus the resulting average is near zero.

\begin{figure}[tb!]
	\centering
	\subfloat[Ground Truth]{
		\includegraphics[scale=.5]{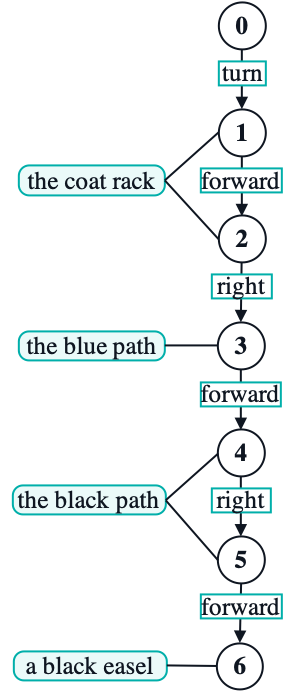}
	}
	\subfloat[Predicted]{
		\includegraphics[scale=.5]{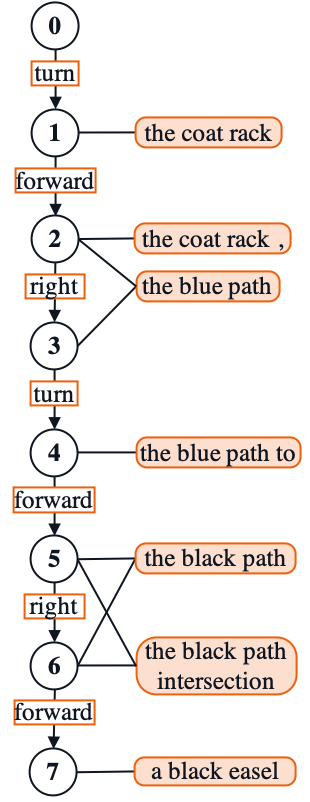}
	}
	\caption{Ground truth (aqua) and predicted (orange) landmark graphs for SAIL route: ``Go towards the coat rack. At the coat rack, take a right onto the blue path. Follow the blue path to the black path intersection and go right onto the black path. Go all the way down until you get to a black easel."}
	\label{fig:sail_example}
\end{figure}

\begin{figure*}[t!]
	\centering
	\subfloat[Ground Truth]{
		\includegraphics[scale=.5]{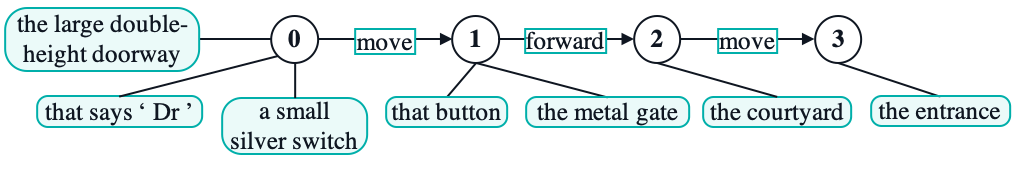}
	}
	\subfloat[Predicted]{
		\includegraphics[scale=.5]{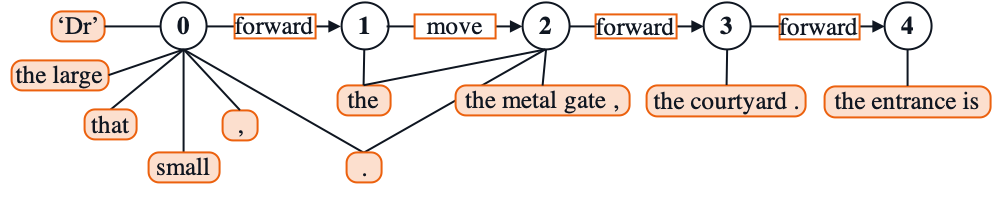}
	}
	\caption{Ground truth (aqua) and predicted (orange) landmark graphs for Delivery route: ``To gain access look for the large double-height doorway, and on the right hand side there's a small silver switch that says `Dr'. Press that button and then push the metal gate, to walk through into the courtyard. The entrance is on the right hand side."}
	\label{fig:lmd_example}
\end{figure*}
\subsection{Landmark Graphs}
Though the decomposition of spatial representation prediction into states and actions is useful both for the model and for indirect evaluation purposes, our goal is to extract complete spatial representations from route instructions. Therefore, we compare predicted landmark graph $g^p$ against the ground truth graph $g^g$ using a modification of graph edit distance \cite{abu2015exact}. This is defined in Equation~\ref{eq:graph_sim}
\begin{equation}
\text{sim}(g^p, g^g) = 1 - \frac{1}{\vert g^p \vert + \vert g^g \vert} \left( \min\limits_{\gamma \in \Gamma(g^p, g^g)}  \sum_{i} c(\gamma_i) \right)
\label{eq:graph_sim}
\end{equation}
where $\vert g \vert$ refers to the sum of the number of edges and number of nodes in graph $g$, $\Gamma(g^p, g^g)$ refers to the set of possible edit paths transforming $g^p$ to $g^g$, $\gamma_i$ is an edit operation in path $\gamma$, and $c$ is the cost of that operation. 

In our strict measure, sim, the insertion and deletion operations have a cost of 1. Substitution costs 0 if the attributes of the two nodes are the same (i.e. an edge labeled with an action can freely replace an edge with the same label, and a landmark node can be substituted for a landmark node with the exact same string), and otherwise costs 1. This measure thus corresponds similarity: the percentage of possible edits that were not necessary in transforming $g^p$ to $g^g$. 

Our more permissive measure, $\text{sim}^{\ell}(g^p, g^g)$, sets the landmark node substitution cost at 1 - $\jaccard({\landmark}^p, {\landmark}^g)$, allowing graphs to be penalized less for inexact landmark matches.

In order to make the comparison to the MLA baseline as fair as possible, we use a by-sentence mapping\footnote{The entities in the MLA code differ slightly from those in SAIL proper; we therefore constructed a new mapping in the same way we did for SAIL.} to replace ground truth entities with surface realizations. Note, however, that the MLA approach is aware of all ground truth entities for each step, not just referents for spans in the instruction; this means their graphs tend to overgenerate for our setting, leading to poorer performance.

Our graph similarity results are shown in Table~\ref{tbl:graph_sim}, where MLA's sim=$\text{sim}^{\ell}$, since inexact matches aren't possible.

As in our other measures, our SAIL model outperforms the MLA baseline while the performance of the Delivery model is significantly lower. Most importantly, though, this measure shows that the spatial representations predicted by these models have a high degree of similarity with the ground truth graphs, particularly when span approximation, rather than exact matching, is incorporated into the measure.

\subsection{Examples}

\textbf{SAIL} An example predicted landmark graph from our SAIL model is shown in Figure~\ref{fig:sail_example}. These results are nearly perfect, though there's an additional \emph{turn} after the \emph{right} (along with a corresponding decision point), and the predicted landmark spans are (at worst) minor variations on the ground truth. 

One important limitation is the absence of landmark coreference resolution: since landmarks must be found anew at each step, the prediction of ``the coat rack ," at step 2 is interpreted as a distinct span from ``the coat rack". Another interesting phenomena is the prediction of ``the black path intersection" and its association with the same decision points as ``the black path". While this is strictly incorrect, a valid interpretation of the route instruction may include this landmark, meaning that the model discovered a useful span that was missed during annotation.

\textbf{Delivery} Figure~\ref{fig:lmd_example} shows a predicted from graph from our Delivery model. Here, the model attended to the parts of the sentence referring to landmarks (including the gate and the courtyard) while also capturing the appropriate spatial relations as expressed by actions (neither the predicted nor the ground truth graph have any turns). 

However, this example demonstrates limitations in both our task framing and our models' capabilities. Conceptually, spatial relations are semantically fuzzy (e.g. two \emph{forward} actions may refer to different real-world distances). In the context of these graphs, this means that while the represented spatial relationships are very similar (and would be interpreted as such by a person) it is still difficult to measure this similarity automatically or for a robot to interpret these actions. Where landmarks are concerned, it's clear that while spans are found in roughly the right locations at the right steps, finding exact span boundaries is difficult for the model: ``the large" and ``small" are missing the most important tokens in their spans, while ``the metal gate ," and ``the entrance is" contain spurious tokens. 

Overall, these predicted graphs confirm that our approach is producing a fairly accurate representations of spaces described by route instructions, but more work is needed.

\section{Conclusion}
We have introduced the task of landmark graph generation and an approach to create these spatial representations by jointly predicting landmark spans and traversal actions. We show that our models yield good performance according to the graph similarity measure we introduce, as well as the related action prediction evaluation measures borrowed from the robotic navigation literature.

However, it is also clear from our results that this work is a first step in need of refinement. Landmark span detection in particular suffers from the simplicity of our approach, and future work will likely incorporate insights from the grounding and coreference resolution literature (particularly approaches like Lee et al. \shortcite{lee2017end}).

We believe that the landmark graph generation task to be a critical next step in the development of landmark-centric navigation technologies, and the results of our Delivery model point to the complexity of the real-world domain and the need for large datasets that capture this complexity.

\bibliography{cervantes}
\bibliographystyle{aaai}
\end{document}